\documentclass[conference]{IEEEtran}
\IEEEoverridecommandlockouts

\usepackage{cite}
\usepackage{amsmath,amssymb,amsfonts}
\usepackage{algorithmic}
\usepackage{graphicx}
\usepackage{textcomp}
\usepackage{xcolor}
\usepackage{bbm}
\usepackage{CJK}
\def\BibTeX{{\rm B\kern-.05em{\sc i\kern-.025em b}\kern-.08em
    T\kern-.1667em\lower.7ex\hbox{E}\kern-.125emX}}
\begin{document}
\twocolumn[
\begin{@twocolumnfalse}
\Huge {IEEE copyright notice} \\ \\
\large {\copyright\ 2024 IEEE. Personal use of this material is permitted. Permission from IEEE must be obtained for all other uses, in any current or future media, including reprinting/republishing this material for advertising or promotional purposes, creating new collective works, for resale or redistribution to servers or lists, or reuse of any copyrighted component of this work in other works.} \\ \\



\vspace{2cm}

\end{@twocolumnfalse}
]
\title{ECG-guided individual identification via PPG}

\author{\IEEEauthorblockN{Riling Wei}
\IEEEauthorblockA{\textit{Zhejiang Lab}\\
Hangzhou, China \\
weirl@zhejianglab.org}
\and
\IEEEauthorblockN{Hanjie Chen}
\IEEEauthorblockA{\textit{Hong Kong Centre for Cerebro-}\\
\textit{-cardiovascular Health Engineering (COCHE)} \\
Hong Kong, China \\
hjchen@hkcoche.org}
\and
\IEEEauthorblockN{Kelu Yao}
\IEEEauthorblockA{\textsuperscript{1}\textit{Zhejiang Lab}\\
\textsuperscript{2}\textit{Zhejiang University}\\
Hangzhou, China \\
yaokelu@zhejianglab.org}
\and
\IEEEauthorblockN{Chuanguang Yang}
\IEEEauthorblockA{\textit{Institute of Computing Technology,} \\
\textit{Chinese Academy of Sciences}\\
Beijing, China \\
yangchuanguang@ict.ac.cn}
\and
\IEEEauthorblockN{Jun Wang\textsuperscript{\dag}\thanks{\textsuperscript{\dag} Corresponding Author.}}
\IEEEauthorblockA{\textit{Zhejiang Lab}\\
Hangzhou, China \\
wangjun@zhejianglab.org}
\and
\IEEEauthorblockN{Chao Li\textsuperscript{\dag}}
\IEEEauthorblockA{\textit{Zhejiang Lab}\\
Hangzhou, China \\
lichao@zhejianglab.org}
}

\maketitle

\begin{abstract}
Photoplethsmography (PPG)-based individual identification aiming at recognizing humans via intrinsic cardiovascular activities has raised extensive attention due to its high security and resistance to mimicry. However, this kind of technology witnesses unpromising results due to the limitation of low information density. To this end, electrocardiogram (ECG) signals have been introduced as a novel modality to enhance the density of input information. Specifically, a novel cross-modal knowledge distillation framework is implemented to propagate discriminate knowledge from ECG modality to PPG modality without incurring additional computational demands at the inference phase. Furthermore, to ensure efficient knowledge propagation, Contrastive Language–Image Pre-training (CLIP)-based knowledge alignment and cross-knowledge assessment modules are proposed respectively. Comprehensive experiments are conducted and results show our framework outperforms the baseline model with the improvement of 2.8\% and 3.0\% in terms of overall accuracy on seen- and unseen individual recognitions.
\end{abstract}


\section{Introduction}
Physiological signal-based individual recognition has emerged and received great attention in recent years. The main idea of such technology is to identify humans via their intrinsic information extracted from heart and brain activities by analyzing PPG~\cite{2020Stochastic}, ECG~\cite{Odinaka},
Electroencephalogram (EEG)~\cite{nakamura2017ear} and etc. As one of the common physiological signals, PPG which reflects human cardiovascular activities has been widely used in the clinic and daily health monitoring due to the convenience and easy to be collected~\cite{sweeney2012artifact}. In recent years, the application of PPG-based biometric technology has emerged~\cite{26karimian2017human}~\cite{28ye2021ppg}~\cite{wei2024pulseid}. On the one hand, with the development of biometric spoofing technology, the security of appearance-based methods, e.g., face recognition~\cite{schroff2015facenet} tend to be damaged while PPG-based methods extract intrinsic information which makes PPG difficult to forge by imposters. On the other hand, investigating the discrimination of PPG signals provides a solid foundation for remote Photoplethsmography(rPPG)~\cite{xu2023ivrr} collected from videos in a non-contact manner, which can be potential biometric traits.
\begin{figure}[t]
	\centering
	\includegraphics[width=7.0cm]{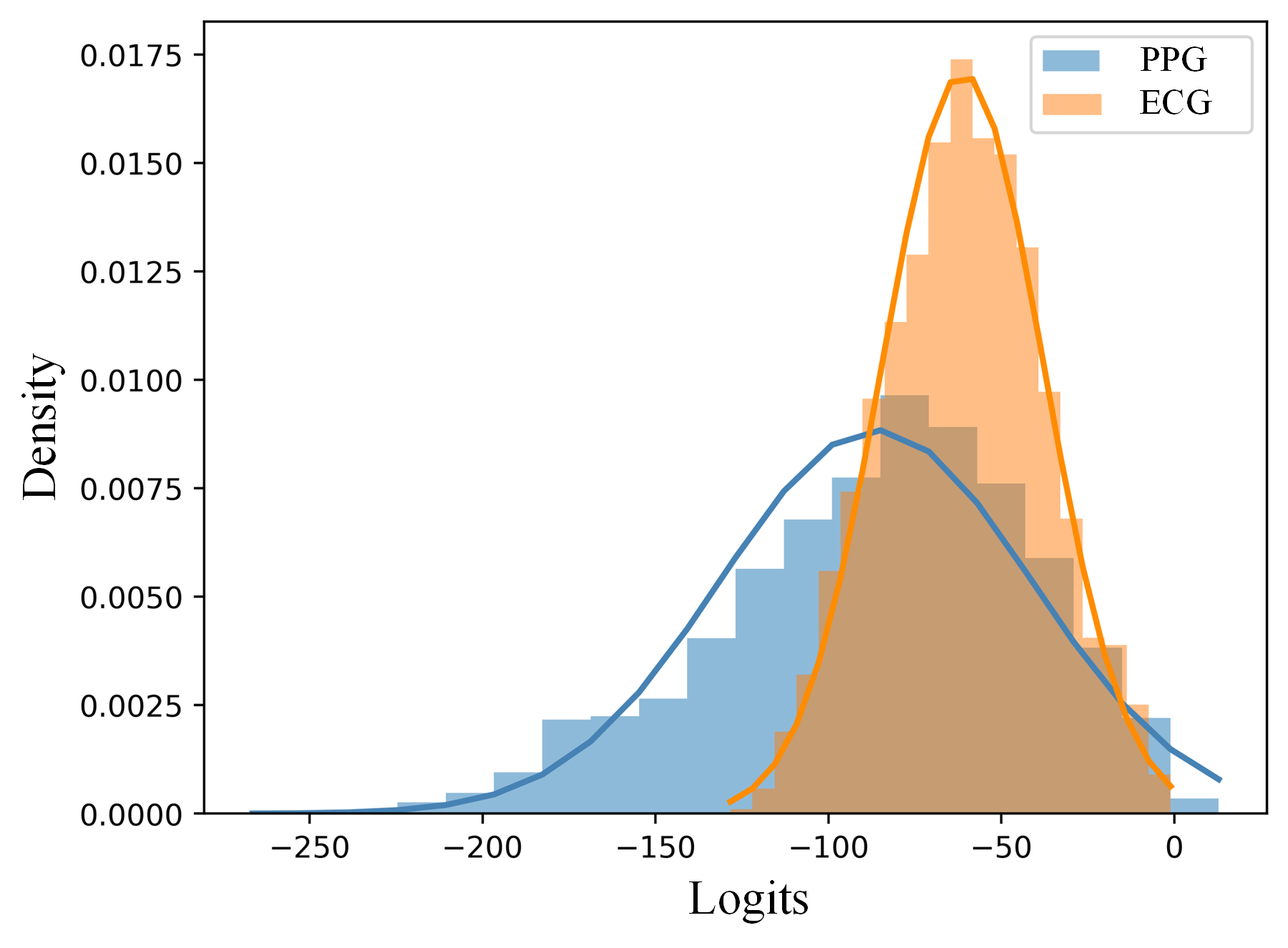}
	\caption{Visualizing the distribution of ECG modalities (the orange part) and PPG modalities (the blue part).}
	\label{fig1}
\end{figure}

However, PPG-based individual recognition usually shows poor performance compared to the traditional biometric approach, especially on very large databases. The main reason might be lower information density~\cite{sarkar2024xkd} caused by the single sensor in the collection step. Employing more modalities to increase information density might be a reasonable manner, e.g., multi-modal learning~\cite{huang2021makes}~\cite{xiao2024neuro}. However, multi-modal learning requires high computation costs and data demands at both the training and inference stages, which tends to be challenging to adopt in real-world scenarios. To tackle the above problems, ECG signal is proposed as a teacher modality to transfer knowledge to PPG signals via a novel cross-modal knowledge distillation framework which enables PPG signals only to be used at the inference phase. The reason for selecting ECG signals can be summarized as: 1) The collection of ECG signals involves more sensors and electrodes which means it contains higher spatial resolutions than PPG signals. As shown in \textbf{Fig.~\ref{fig1}}, the output logits of a model taking ECG and PPG modalities respectively as input on classification tasks are visualized. It can be observed that ECG modality has a sharper distribution compared to PPG, which indicates ECG signals can provide better supervision for PPG signals~\cite{sarkar2024xkd}. 2) Compared to other discriminative modalities, e.g., face, voice, etc., ECG signals bridge a reasonable modality gap with PPG signals because both ECG and PPG signals reflect humans' cardiovascular representations. The contribution of this paper can be summarized as follows:




\begin{itemize}
\item To the best of our knowledge, we are the first to propose ECG signal as a new modality to teach PPG signal to learn to discriminate distribution.
\item To ensure highly efficient knowledge propagation, we propose a plug-and-play CLIP-based knowledge alignment module to project relation-based knowledge to a common latent space regardless of extra computational cost at the inference stage.
\item We proposed a cross-knowledge assessment module to evaluate teaching and learning results.
\item We conduct comprehensive experiments on different networks to valid the generalization capability of the proposed method.
\end{itemize}

\section{Related Works}
\textbf{PPG-based individual identification}. The main idea is to distinguish humans in large databases, which are regarded as multi-class classification tasks. Handcrafted-based method~\cite{26karimian2017human}~\cite{28ye2021ppg} is proposed which tends to be effective on small-scale databases while failing on large-scale databases due to unpromising generalization capability. To tackle this issue, a novel end-to-end deep learning-based method has been proposed in recent years and shows promising results thanks to superior generalization capability~\cite{wei2024pulseid}.

\textbf{Cross-modal knowledge distillation (CMKD)}. Knowledge distillation (KD)~\cite{hinton2015distilling} aims at transferring knowledge from a large model to a smaller one to optimize memory cost and inference speed and was widely used in computer vision~\cite{yang2022cross}~\cite{yang2022mcl}~\cite{yang2023okd} and cross-modal learning tasks~\cite{yang2024clip}. Unlike conventional KD, which accepts the same input, in CMKD, the teacher and the student use different modalities as input. The goal of CMKD is to propagate knowledge from a discriminate modality to a weaker one. For instance, Jin \textit{et al.}~\cite{jin2023cross} proposed CMKD to improve the performance of speaker recognition supervised by facial representations.
\section{Methodology}
\textbf{Notations}. Given a teacher model $\mathcal{T}$ and a student model $\mathcal{S}$, which take ECG signal and PPG signal as input respectively. We let $\mathcal{T}_{f}$ and $\mathcal{S}_{f}$ be the feature extractor of $\mathcal{T}$ and $\mathcal{S}$ respectively.  $\mathcal{T}_{cls}$ and $\mathcal{S}_{cls}$ are the classifier of $\mathcal{T}$ and $\mathcal{S}$. $\mathcal{F}_{t}$ and $\mathcal{F}_{s}$ are the feature embedding of $\mathcal{T}_{f}$ and $\mathcal{S}_{f}$.
\begin{figure}[h]
	\centering
	\includegraphics[width=9.0cm]{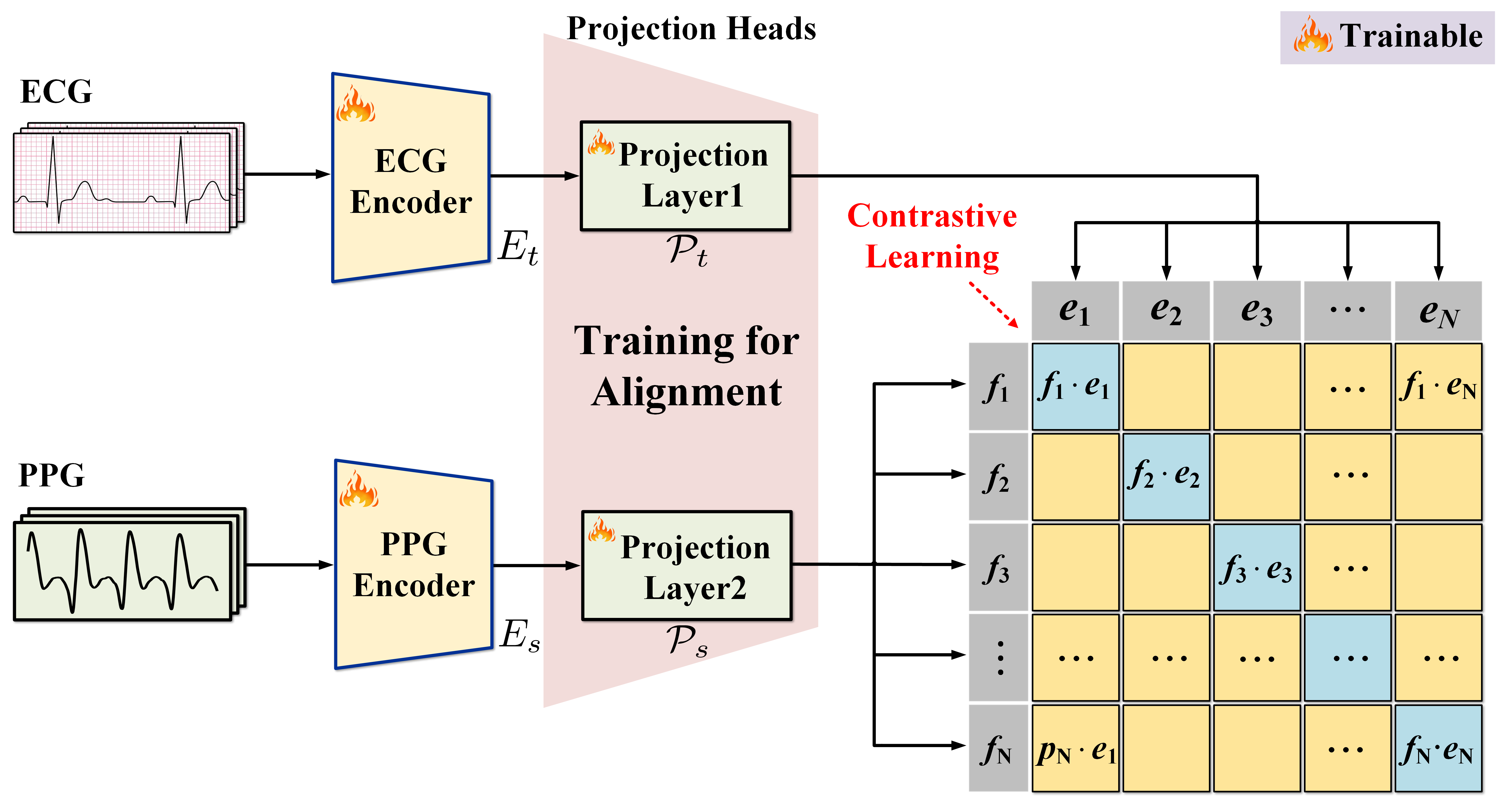}
	\caption{The training stage of CLIP-based knowledge alignment module.}
	\label{fig2}
\end{figure}
\subsection{CLIP-based Knowledge Alignment (CLIP)}
In CMKD studies, one of the biggest challenges is the domain gap between different modalities which might damage the efficiency of knowledge propagation. Hence, in this part, we design a plug-and-play knowledge alignment module which is used to project $\mathcal{F}_{t}$ and $\mathcal{F}_{s}$ to a common feature space before transferring knowledge. As it is only used in the training phase, there are no further computation requirements at the inference stage. Specifically, we first train an alignment model containing 2 projection layers in a CLIP manner~\cite{radford2021learning}. The overall architecture consists of two encoders, dubbed ${E}_{t}$ and ${E}_{s}$ respectively, and two projection heads, noted as $\mathcal{P}_{t}$ and $\mathcal{P}_{s}$, illustrated in \textbf{Fig.~\ref{fig2}}. As an example, ResNet34 and 1 Fully Connected layer are employed as encoders and projection heads respectively. The dimension of embeddings are projected from 512 dimensions to 256 dimensions. InfoNCE~\cite{gutmann2010noise} is adopted to train the model which is implemented as follows:
\begin{equation}
	\label {eq1}
	\resizebox{0.99\hsize}{!}{$ 
		\ \mathcal{L}_{NCE} = \frac{1}{2N} (\sum_{i=1}^{N} -log(\frac {{e}^{{{s}_{i,j}}/{\tau}}}{\sum_{i=1}^{N}{e}^{{s}_{i,j}/{\tau}}}) +\sum_{i=1}^{N} -log(\frac {{e}^{{{s}_{i,i}}/{\tau}}}{\sum_{j=1}^{N}{e}^{{s}_{j,i}/{\tau}}}))
		$}
\end {equation}

Where, ${s}_{j,i}$ is the dot operation between $i$ th ECG embedding ${e}_{i}$ and $j$ th PPG embedding ${f}_{j}$. $\tau$ is temperature. 
It should be noted that only $\mathcal{P}_{t}$ and $\mathcal{P}_{s}$ are utilized in the next step, which projects $\mathcal{F}_{t}$ and $\mathcal{F}_{s}$ to a common latent space respectively. The aligned embedding noted as $\mathcal{F}_{t}^{a}$ and $\mathcal{F}_{s}^{a}$. Subsequently, Maximum Mean Discrepancy (MMD)~\cite{gretton2006kernel} and Triplet Loss~\cite{schroff2015facenet} are employed as relation-based knowledge, noted as $\mathcal{L}_{rel}$, on seen- and unseen- individual recognition testings respectively. MMD can be summarized as:
\begin{equation}
	\label {eq2}
	\ \mathcal{L}_{mmd}({\mathbbm{P}}_{t},{\mathbbm{P}}_{s}) = {||{\mathbbm{E}}_{{x}_{t}\sim{\mathbbm{P}}_{t}}[k(\cdot,{x}_{t})]-{\mathbbm{E}}_{{x}_{s}\sim{\mathbbm{P}}_{s}}[k(\cdot,{x}_{s})]||}_{{\mathcal{H}}_{k}}
\end {equation}
Here, ${\mathbbm{P}}_{t},{\mathbbm{P}}_{s}$ are regarded as the feature distribution of $\mathcal{T}_{f}$ and $\mathcal{S}_{f}$. ${x}_{t}$ and ${x}_{s}$ are the sample of ${\mathbbm{P}}_{t}$ and ${\mathbbm{P}}_{s}$. ${||\cdot||}_{{\mathcal{H}}_{k}}$ is the RKHS norm. $k$ is the Gaussian kernal with bandwidth $\sigma$ summarized as:
\begin{equation}
	\label {eq3}
	\ k({x}_{t},{x}_{s}) = exp(\frac{-{||{x}_{t}-{x}_{s}||}^{2}}{2{\sigma}^{2}})
\end {equation}
The Triplet loss is summarized as:
\begin{equation}
	\label {eq4}
	\ \mathcal{L}_{triplet}({\mathbbm{P}}_{s},{\mathbbm{P}}_{t}^{+},{\mathbbm{P}}_{t}^{-}) = max(d({\mathbbm{P}}_{s},{\mathbbm{P}}_{t}^{+})-d({\mathbbm{P}}_{s},{\mathbbm{P}}_{t}^{-})+m, 0)
\end {equation}
Here, ${\mathbbm{P}}_{t}^{+}$ and ${\mathbbm{P}}_{t}^{-}$ are the teacher embedding extracted from the same and different subjects with ${\mathbbm{P}}_{s}$. $d(\cdot)$ is the distance measurement. $m$ is a hyperparameter, which is set to 1.
\begin{figure}[t]
	\centering
	\includegraphics[width=9.0cm]{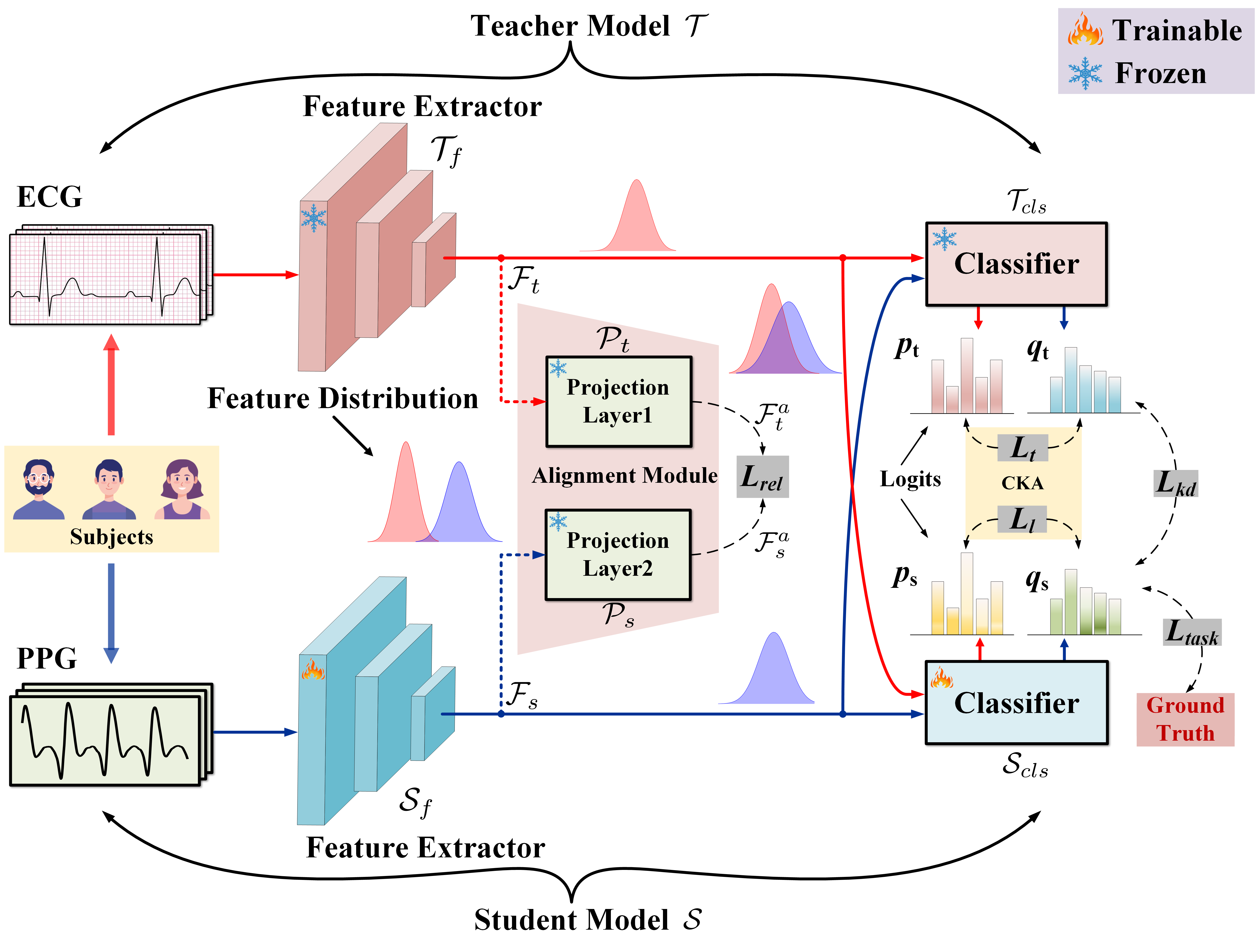}
	\caption{The framework of the proposed methods.}
	\label{fig3}
\end{figure}

\subsection{Cross-Knowledge Assessment (CKA)}
We first implemented conventional KD proposed by~\cite{hinton2015distilling} as follows:
\begin{equation}
	\label {eq5}
	\ \mathcal{L}_{kd}= H({p}_{t}, {q}_{s};\tau)
\end {equation}
Where $\tau$ is the temperature, that is used to balance the smoothness of output. In our experiments, $\tau$ is set to 4. ${p}_{t}$ and ${q}_{s}$ are the output of $\mathcal{T}_{cls}$ and $\mathcal{S}_{cls}$. $H(\cdot)$ indicates Kullback Leibler (KL) divergence.

In this part, CKA is proposed to evaluate the learning and teaching outcomes by the teacher and student respectively as illustrated in \textbf{Fig.~\ref{fig3}}. To achieve this goal, $\mathcal{F}_{s}$ are fed into $\mathcal{T}_{cls}$ and obtain ${q}_{t}$. $\mathcal{T}_{cls}$ is regarded as a referee to distinguish how much knowledge has been learned by $\mathcal{S}$ by measuring the similarity of ${p}_{t}$ and ${q}_{t}$. The discrepancy of the knowledge between $\mathcal{T}$ and $\mathcal{S}$ is implemented as:
\begin{equation}
	\label {eq6}
	\ \mathcal{L}_{t}= H({p}_{t}, {q}_{t};\tau)
\end {equation}
\begin{equation}
	\label {eq7}
	\ \mathcal{L}_{l}= H({p}_{s}, {q}_{s};\tau)
\end {equation}
Here, ${p}_{s}$ is the output of $\mathcal{S}_{cls}$ which takes $\mathcal{F}_{t}$ as input. Thus, the total cross-knowledge assessment loss can be written as:
\begin{equation}
	\label {eq8}
	\ \mathcal{L}_{cross-kd}= \frac{1}{2}(\mathcal{L}_{t}+\mathcal{L}_{l})
\end {equation}
Finally, the overall loss function can be implemented as:
\begin{equation}
	\label {eq9}
	\ \mathcal{L}_{all}= \mathcal{L}_{task}+\mathcal{L}_{kd}+\mathcal{L}_{rel}+\mathcal{L}_{cross-kd}
\end {equation}
where $\mathcal{L}_{task}$ is the task-related loss function. Specifically, in classification task $\mathcal{L}_{task}$ is regarded as cross-entropy loss while in retrieval tasks it is known as a bank of cross-entropy loss and triplet loss.
	
\section{Experimental Setup}
\begin{table}
	\setlength{\tabcolsep}{2pt}
	\renewcommand{\arraystretch}{1.2}
	\begin{center}
		\caption{Performance of sample-wise testing compared to baseline approach.}
		\label{tab1}
		\resizebox{\linewidth}{!}{
			\begin{tabular}{ l  c  c c c c c}
				\hline
				\textbf{Model} & \multicolumn{2}{c}{\textbf{ResNet34}} & \multicolumn{2}{c}{\textbf{MobileNetV1}} & \multicolumn{2}{c}{\textbf{ShuffleNetV1}}\\
				\hline
				\textbf{Method} & \textbf{OA(\%)$\uparrow$} & \textbf{F1(\%)$\uparrow$} & \textbf{OA(\%)$\uparrow$} & \textbf{F1(\%)$\uparrow$} & \textbf{OA(\%)$\uparrow$} & \textbf{F1(\%)$\uparrow$}\\
				\hline
				Baseline & 93.5 & 92.9 & 93.0 & 92.3 & 87.9 & 86.8\\
				
				\textbf{Ours} & ${95.1}^{\textcolor{red}{\uparrow1.6}}$ & ${94.5}^{\textcolor{red}{\uparrow1.6}}$ & ${94.1}^{\textcolor{red}{\uparrow1.1}}$ & ${93.5}^{\textcolor{red}{\uparrow1.2}}$ & ${90.7}^{\textcolor{red}{\uparrow2.8}}$ & ${89.9}^{\textcolor{red}{\uparrow3.1}}$\\
				
				Teacher & 96.4 & 96.1 & 95.3 & 94.8 & 92.4 & 91.7\\
				\hline
			\end{tabular}
		}
	\end{center}
\end{table}
\begin{table}
	\vspace{-1.0em}
	\setlength{\tabcolsep}{2pt}
	\renewcommand{\arraystretch}{1.2}
	\begin{center}
		\caption{Performance of subject-wise testing compared to baseline approach on ResNet34.}
		\label{tab2}
		\resizebox{\linewidth}{!}{
			\begin{tabular}{ l  c  c c c c c}
				\hline
				\textbf{N-Shot} & \multicolumn{2}{c}{\textbf{1-Shot}} & \multicolumn{2}{c}{\textbf{5-Shot}} & \multicolumn{2}{c}{\textbf{10-Shot}}\\
				\hline
				\textbf{Method} & \textbf{OA(\%)$\uparrow$} & \textbf{EER(\%)$\downarrow$} & \textbf{OA(\%)$\uparrow$} & \textbf{EER(\%)$\downarrow$} & \textbf{OA(\%)$\uparrow$} & \textbf{EER(\%)$\downarrow$}\\
				\hline
				Baseline & 83.1 & 4.0 & 91.1 & 2.6 & 91.0 & 2.9\\
				
				\textbf{Ours} & ${86.1}^{\textcolor{red}{\uparrow3.0}}$ & ${2.2}^{\textcolor{red}{\downarrow1.8}}$ & ${92.0}^{\textcolor{red}{\uparrow0.9}}$ & ${2.0}^{\textcolor{red}{\downarrow0.6}}$ & ${91.4}^{\textcolor{red}{\uparrow0.4}}$ & ${1.9}^{\textcolor{red}{\downarrow1.0}}$\\
				
				Teacher & 92.4 & 1.1 & 93.9 & 1.2 & 94.5 & 1.1\\
				\hline
			\end{tabular}
		}
	\end{center}
\end{table}
\begin{table}[h]
	\setlength{\tabcolsep}{4.5pt}
	\renewcommand{\arraystretch}{1.2}
	\begin{center}
		\caption{Performance of sample-wise testing on MIMIC dataset compared to state-of-the-art methods.}
		\label{tab3}
		\begin{tabular}{ l  c  c c c c c}
			\hline
			\textbf{Model} & \multicolumn{2}{c}{\textbf{ResNet34}} & \multicolumn{2}{c}{\textbf{MobileNetV1}} & \multicolumn{2}{c}{\textbf{ShuffleNetV1}}\\
			\hline
			\textbf{Method} & \textbf{OA(\%)} & \textbf{F1(\%)} & \textbf{OA(\%)} & \textbf{F1(\%)} & \textbf{OA(\%)} & \textbf{F1(\%)}\\
			\hline
			KD\cite{hinton2015distilling} & \underline{94.7} & \underline{94.2} & \underline{93.9} & \underline{93.4} & 89.0 & 88.3\\
			
			RKD\cite{park2019relational} & 94.0 & 93.5 & 92.7 & 92.0 & 88.1 & 87.4\\
			
			DKD\cite{zhao2022decoupled} & 94.1 & 93.6 & 93.4 & 92.9 & 85.3 & 84.1\\
			
			LKD\cite{ba2014deep} & 93.9 & 93.4 & 93.6 & 92.9 & \underline{89.4} & \underline{88.6}\\
			
			CTKD\cite{li2023curriculum} & 94.5 & 93.9 & 93.7 & 93.0 & 88.8 & 87.9 \\
			
			STKD\cite{sun2024logit} & 94.6 & \underline{94.2} & \underline{93.9} & 93.2 & 88.9 & 88.2\\ 
			
			\textbf{Ours} & \textbf{95.1} & \textbf{94.5} & \textbf{94.1} & \textbf{93.5} & \textbf{90.7} & \textbf{89.9}\\ 
			\hline
		\end{tabular}
	\end{center}
\end{table}
\begin{table}[h]
	\vspace{-1.0em}
	\setlength{\tabcolsep}{14pt}
	\renewcommand{\arraystretch}{1.4}
	\begin{center}
		\caption{Ablation study of the proposed framework on sample-wise testing.}
		\label{tab4}
		\begin{tabular}{ c  c  c c c}
			\hline
			\textbf{KD} & \textbf{CLIP} & \textbf{CKA} & \textbf{OA(\%)} & \textbf{F1(\%)}\\
			\hline
			$\checkmark$ &  &  & 94.7 & 94.2\\
			
			$\checkmark$ & $\checkmark$ &  & 94.8 & 94.3\\
			
			$\checkmark$ & & $\checkmark$ & 95.0 & 94.4\\
			
			$\checkmark$ & $\checkmark$ & $\checkmark$ & \textbf{95.1} & \textbf{94.5}\\
			\hline
		\end{tabular}
	\end{center}
\end{table}
\subsection{Dataset}
MIMIC dataset~\cite{goldberger2000physiobank} is employed which contains 943 subjects and 12000 records for both ECG and PPG signals. All signals are sampled as 125Hz. We only use recordings longer than 8 minutes to evaluate our proposed framework. In our work, 341 individuals with high-quality ECG signals and PPG signals are involved.

\subsection{Experiments and Evaluation Metrics}
A 300-point sliding window is utilized to segment signal samples from original signals into non-overlapping samples for both ECG and PPG signals. Subsequently, data standardization is employed to remove noise. In this work, we utilize 1D ResNet34~\cite{he2016deep}, MobileNetV1~\cite{2017MobileNets}, and ShuffleNetV1~\cite{zhang2018shufflenet} to evaluate the generalization capability of the proposed framework. Heterogeneous model architectures might introduce lower similarity between the teacher and the student compared to homogeneous models~\cite{hao2023ofa}. To focus on our main goal, we just investigate homogeneous models of teacher and student. We conduct two sets of experiments, namely:

1. \textbf{Sample-wise individual recognition}: Aiming at recognizing humans in the seen domain. 341 subjects are involved in this testing. For the training set, we use the first 6.4 minutes recordings of each subject. For the testing set, the rest 1.6 minutes of recordings are utilized for evaluation.

2. \textbf{Subject-wise individual recognition}: The goal of this testing is to evaluate the generalization capability of the proposed framework. Specifically, 272 subjects are involved in the training set and the other 69 subjects are selected randomly in the testing phase. In other words, the test individuals are disjoint from the training individuals.

On sample-wise testing, we utilize \textbf{O}verall \textbf{A}ccuracy (\textbf{OA}) and \textbf{F1}-score(\textbf{F1}) for classification tasks. On subject-wise testing, \textbf{OA} and \textbf{E}qual \textbf{E}rror \textbf{R}ate (\textbf{EER}) are employed.
\section{Results and Discussion}
\subsection{Sample-wise Testing}
\label{ssec:Sample-wise}
We first conduct experiments on sample-wise recognition, which is used to identify individuals in the seen domain. Firstly, we trained a baseline model supervised by cross-entropy loss without KD on MIMIC datasets for classification tasks as shown in \textbf{Tab.~\ref{tab1}}. The teacher model takes ECG as input under the supervision of only cross-entropy loss as well. From \textbf{Tab.~\ref{tab1}}, it can be seen that employing ECG signals can significantly improve recognition performance regardless of different architectures. For example, our framework boosts the performance of ShuffleNetV1 by 2.8\% and 3.1\% in terms of \textbf{OA} and \textbf{F1}.
\subsection{Subject-wise Testing}
Experiments on recognizing subjects in the unseen domain are conducted on ResNet34. The baseline model and teacher model are supervised by a bank of cross-entropy loss and triplet loss. We adopt a N-shot manner to test our framework. Specifically, N is set to 1, 5, and 10 respectively and the result is shown in \textbf{Tab.~\ref{tab2}}. From the table, it can be seen that our framework achieves superior performance with the improvement of 0.4\% - 3.0\% and 0.6\% - 1.8\% in terms of \textbf{OA} and \textbf{EER} compared to the baseline model, which indicates the promising generalization capability of the proposed method.
\subsection{Compare with SOTA methods}
We conduct experiments to evaluate the performance of our framework by comparing it to state-of-the-art approaches in knowledge distillation. \textbf{Tab.~\ref{tab3}} illustrates that the proposed method achieves state-of-the-art performance among several mainstream KD methods, which indicates that our approach can propagate knowledge effectively between different modalities.

\subsection{Ablation studies}
We conduct the following ablation testing to evaluate the effectiveness of the proposed CLIP module and CKA module with ResNet34. It can be seen clearly from \textbf{Tab.~\ref{tab4}} that only utilizing a CLIP module or CKA can boost the performance compared to the baseline. Besides, applying both CLIP and CKA achieves the best performance which is 95.1\% and 94.5\% in terms of \textbf{OA} and \textbf{F1}.

\subsection{Visualization}
\begin{figure}[t]
	\centering
	\includegraphics[width=8.0cm]{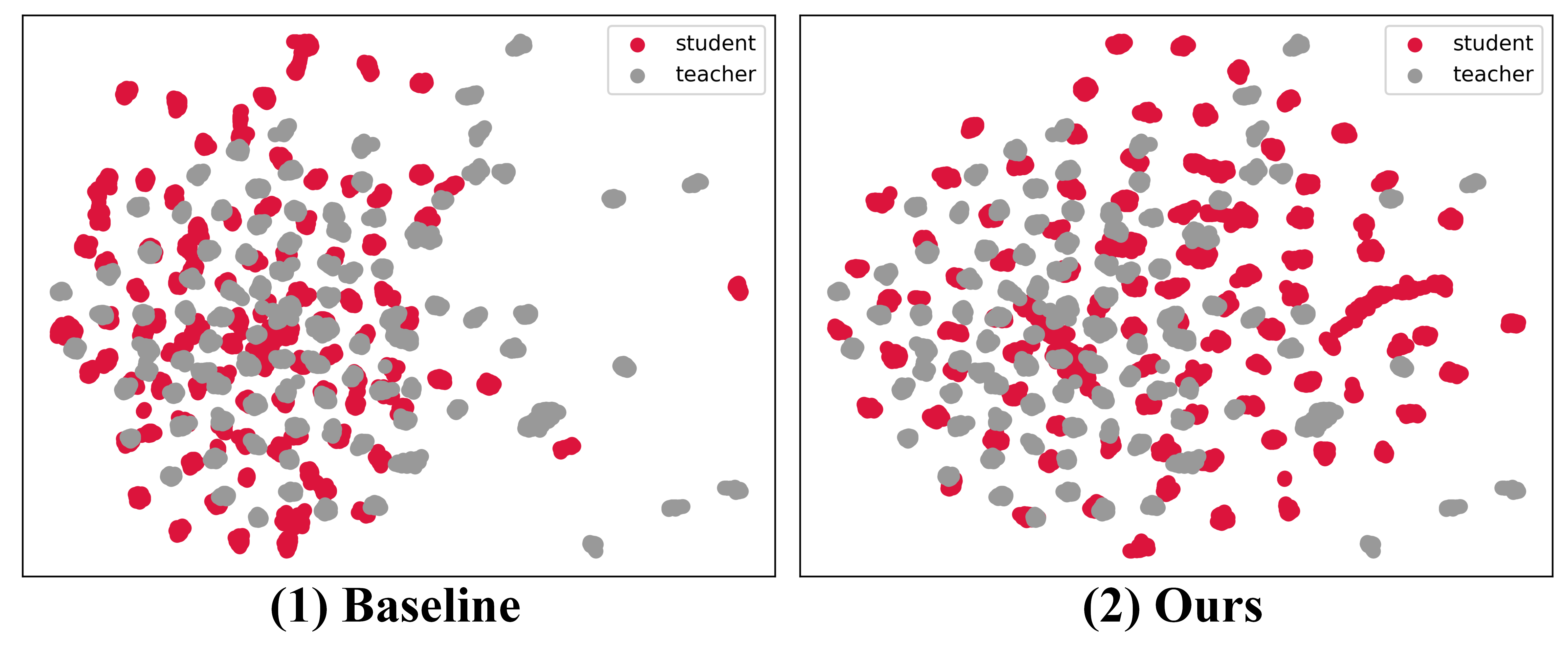}
	\caption{The visualization of feature distribution in sample-wise testing. The red and grey parts denote student and teacher representations respectively. (1) shows that representations of the baseline. (2) indicates the feature distributions of the proposed framework.}
	\label{fig4}
\end{figure}

As shown in \textbf{Fig.~\ref{fig4}}, T-SNE~\cite{van2008visualizing} is employed to visualize the feature distributions of the teacher and student employing ShuffleNetV1 as the backbone. We observe that the proposed framework shows better inter-class distance compared to the baseline. Hence, identities tend to be easier to distinguish. This also indicates the effectiveness of our method.
\section{Conclusion}
In this study, we introduce a cross-modal knowledge distillation framework designed to enhance the performance of PPG-based individual recognition, guided by ECG modalities. To ensure efficient knowledge propagation, a CLIP-based knowledge alignment module is proposed to project both ECG and PPG modalities into a common latent space. Subsequently, cross-knowledge assessment is conducted to evaluate the outcomes of the teaching and learning processes. Experimental results demonstrate that our approach achieves state-of-the-art performance in both sample-wise and subject-wise testing.

However, we also identify limitations within the proposed framework. Specifically, as the number of N-shots increases from 5 to 10 in subject-wise testing, our framework exhibits performance degradation, contrary to the results observed with the teacher model. We consider these findings as directions for future research.

\section*{Acknowledgment}
This work was supported by Zhejiang Provincial Elite Project (No. 2023C01042).
This work was also supported in part by InnoHK Project at Hong Kong Centre for Cerebro-cardiovascular Health Engineering (COCHE).

\bibliographystyle{IEEEtran}
\bibliography{IEEEabrv,refs}


\end{document}